\documentclass{article}

\usepackage{graphicx} 

\usepackage{arxiv}

\usepackage[utf8]{inputenc} 
\usepackage[T1]{fontenc}    
\usepackage[dvipsnames]{xcolor}

\usepackage{hyperref}       
\usepackage{url}            
\usepackage{booktabs}       
\usepackage{array}
\usepackage{nicefrac}       
\usepackage{microtype}      

\usepackage{amsmath}
\usepackage{amsfonts}
\usepackage{amsthm}
\usepackage{float}
\usepackage{multirow}
\usepackage{subcaption}
\usepackage{rotating}

\usepackage[square,numbers]{natbib}

\usepackage{theoremref}

\usepackage{graphicx}
\usepackage{doi}

\title{Characterizing Satellite Geometry via Accelerated 3D Gaussian Splatting}

\author{ \hspace{1mm}V. Minh Nguyen\\
        NEural TransmissionS (NETS) Lab\\
	Florida Institute of Technology\\
	\href{mailto:vnguyen2014@my.fit.edu}{\texttt{vnguyen2014@my.fit.edu}} \\
     \And
    	\hspace{1mm}Emma Sandidge \\
            NEural TransmissionS (NETS) Lab\\
    	Florida Institute of Technology\\
    	\href{mailto:esandidge2020@my.fit.edu}{\texttt{esandidge2020@my.fit.edu}} \\
     \And
    	\hspace{1mm}Trupti Mahendrakar \\
            Autonomy Lab\\
    	Florida Institute of Technology\\
    	\href{mailto:tmahendrakar2020@my.fit.edu}{\texttt{tmahendrakar2020@my.fit.edu}} \\
	\And
	\href{https://orcid.org/0000-0002-5524-629X}{\includegraphics[scale=0.06]{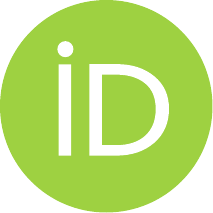}\hspace{1mm}Ryan T.~White} \\
        NEural TransmissionS (NETS) Lab\\
	Florida Institute of Technology\\
	\href{mailto:rwhite@fit.edu}{\texttt{rwhite@fit.edu}} \\
}

\date{}

\usepackage{arydshln}

\begin{document}

\maketitle

\begin{abstract}
    The accelerating deployment of spacecraft in orbit have generated interest in on-orbit servicing (OOS), inspection of spacecraft, and active debris removal (ADR). Such missions require precise rendezvous and proximity operations in the vicinity of non-cooperative, possible unknown, resident space objects. Safety concerns with manned missions and lag times with ground-based control necessitate complete autonomy. This requires robust characterization of the target's geometry. In this article, we present an approach for mapping geometries of satellites on orbit based on 3D Gaussian Splatting that can run on computing resources available on current spaceflight hardware. We demonstrate model training and 3D rendering performance on a hardware-in-the-loop satellite mock-up under several realistic lighting and motion conditions. Our model is shown to be capable of training on-board and rendering higher quality novel views of an unknown satellite nearly 2 orders of magnitude faster than previous NeRF-based algorithms. Such on-board capabilities are critical to enable downstream machine intelligence tasks necessary for autonomous guidance, navigation, and control tasks.
\end{abstract}

\section{Introduction}

The recent reduction in the costs of launching spacecraft has led to an acceleration in the deployment of large satellite constellations and human spaceflight exploration missions. However, retired satellites and collisions of space objects in orbit have resulted in an exponential rise in untracked space debris. This poses collision risks to new launches, which could spread thousands of new pieces of debris, causing catastrophic cascading failures.

Numerous efforts in the field of robotic active debris removal (ADR) and on-orbit servicing (OOS) have been conducted around known cooperative and non-cooperative spacecraft. In 1997, ETS-VII \cite{Oda_2000jrm} by NASDA (now JAXA) made a groundbreaking leap by demonstrating the first ever autonomous docking with a target (cooperative) spacecraft. Following that other technology development missions were conducted around cooperative spacecraft such as NASA, DARPA, AFRL’s DART, XSS-10  \cite{davis_xss-10_2004}, XSS-11 \cite{afrl_xss-11_2011}, ANGELS \cite{afrl_automated_2014}, MiTEX and Orbital Express. In 2020 and 2021, Northrup Grumman conducted the first ever commercial OOS operations with its MEV-1 \cite{intelsat_mev-1_2020} and MEV-2  \cite{rainbow_mev-2_2021, pyrak_performance_2022} satellites to take over station keeping of IS-901 and IS-10-02 respectively. In 2022 as part of demonstrating autonomous capture of uncontrolled object, Astroscale’s ELSA-d mission’s servicer demonstrated autonomous rendezvous with the client prior to pausing the mission due to failed thrusters \cite{forshaw_elsa-d_2019}. All these missions were conducted around known and/or cooperative target objects. However, in reality the RSOs have unknown structure, mass, geometry and attitude. Performing OOS around unknown non-cooperative spacecraft remains an unresolved challenge. 

\begin{figure}[h]
    \centering
    \includegraphics[width=0.5\linewidth]{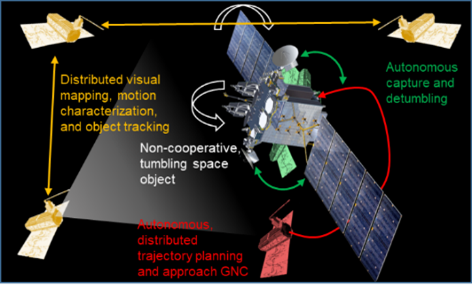}
    \captionsetup{justification=centering}
    \caption{Concept for DSS-based target object characterization, trajectory planning, approach and detumbling  \cite{mahendrakar_real-time_2021}}
    \label{fig:concept}
\end{figure}

To perform OOS and ADR around these unknown RSOs, it is essential to autonomously characterize the geometry, identify points for capture, plan and execute safe approach paths, capture the target, and stabilize its attitude. Distributed Satellite Systems (DSS), groups of small, cooperative satellites each equipped with a system for relative navigation, a robust propulsion system, and a flexible capture mechanism, offer a scalable solution for OOS and ADR tasks involving large non-cooperative targets as the swarm can distribute the forces and moments required to detumble the RSO. A concept of the DSS is illustrated in Figure \ref{fig:concept}.

Working together, these satellite swarms will map the target's geometry (the focus of this article), assess its motion and structural condition, pinpoint possible capture points and collision risks, devise safe approach routes to these points, and collectively plan and carry out maneuvers to stabilize the target's spin. The complexity of these operations, especially the skills needed for mapping the geometry accurately necessitates the application of machine learning technologies. This is crucial to replicate the inherent abilities of humans in these tasks.


The complexity of this approach necessitates complete autonomy during the rendezvous and proximity operation (RPO) phases. Although previous research has shown the integration of AI-based machine vision for navigation \cite{mahendrakar_use_2021} and artificial potential field for guidance \cite{cutler_artificial_2022, mahendrakar_autonomous_2023} as a feasible method for autonomous rendezvous with these unknown RSOs this approach required multiple observers along with multiple chasers in the system. Although theoretically feasible, it is impractical for real-world application, leading to a need for more streamlined and effective solutions in autonomous space servicing missions.


It is much preferred to deploy a single observer that will inspect and fully characterize the geometry and components of the non-cooperative RSO, and then guide the chasers, but this requires full characterization of the geometry and components (e.g., satellites and antennas) of the RSO that can be done using on-board spaceflight hardware. While multi-observer swarms can characterize 3D geometry by triangulating points several observers can view, this is more challenging for a single observer.


This article aims to tackle the first of these two problems: characterizing the geometry of the RSO at low computational cost. This is key in developing enabling software for autonomous RPO with a single observer. 3D Gaussian Splatting \cite{kerbl_3d_2023} will be deployed to characterize the geometry of a target RSO and generate novel viewing angles of it to build a full 3D understanding of the RSO.

The primary contributions of this article include:
\begin{itemize}
    \item An effective low-compute 3D Gaussian Splatting model optimized for 3D reconstruction of an \textit{unknown} RSO capable of deployment on spaceflight hardware.
    \item Hardware-in-the-loop experiments demonstrating 3D rendering performance under realistic lighting and motion conditions.
    \item A comparison of NeRF, D-NeRF, Instant NeRF, 3D Gaussian Splatting, and 4D Gaussian Splatting algorithms in terms of 3D rendering quality, runtimes, and computational costs.
\end{itemize}

\section{Related Work}

\subsection{Computer Vision for On-Orbit Operations}

Traditional computer vision techniques are widely familiar in the field of guidance, navigation and controls and act as the first step to navigation. However, traditional techniques are limited to the presence of known/cooperative features which is not the case for unknown satellites. In recent years, neural network-based computer vision techniques have shown highly promising results for performing in non-cooperative environments and are actively being used for autonomous navigation in robotics and the automobile industry, for example. Extensive research is being performed to identify and track \cite{attzs_comparison_2023} components of non-cooperative known spacecraft \cite{viggh_training_2023} and unknown spacecraft \cite{mahendrakar_use_2021, dung_spacecraft_2021, mahendrakar_performance_2022, faraco2022instance}, such as solar panels, antennas, body and thrusters. Knowing the presence of components and its relative positions could assist with autonomously rendezvousing to the RSO as shown in  \cite{mahendrakar_impact_2023, mahendrakar_autonomous_2023}. However, as described earlier, being able to characterize the RSO with a single observer reduces the risk of collision while improving reliability in the navigation model. 

Research derived from the spacecraft pose estimation challenge based on datasets \cite{Kisantal_2020, park_speed_2022} show that the pose of a known non-cooperative spacecraft can be reliably estimated using a monocular camera \cite{park_towards_2019, sharma_neural_2020, lotti_deep_2023, piazza_monocular_2022}. However, these techniques fail for an unknown spacecraft. 

Photogrammetry techniques based on neural radiance fields (NeRFs) \cite{mildenhall_nerf_2021} and generative radiance fields (GRAFs) \cite{schwarz2020graf} have gained significant popularity to perform 3D reconstruction of an object based on images. In \cite{mergy2021vision}, NeRF and GRAF generated 3D models of SMOS and a CubeSat and showed that NeRF achieved better performance than GRAF. In  \cite{caruso_3d_2023}, the authors further investigated NeRF along with instant NeRF \cite{muller_instant_2022} and dynamic NeRF \cite{pumarola_dnerf_2020} on hardware-in-the-loop (HIL) images of a satellite mockup. It was concluded that instant NeRF, which is capable of running on hardware with spaceflight heritage \cite{kulu_dodona_nodate} is preferred for its fast computational time and relatively good performance. In \cite{park_rapid_2024}, the author takes a unique approach to identify pose of an unknown spacecraft by fitting superquadrics in the mask of a single satellite image. Though this approach effectively identifies high-level details such as solar panels, it lacks smaller crucial details such as apogee kick motors (if present), smaller cylindrical rods or check if a component is still intact that may be a potential docking feature as described in \cite{mahendrakar_autonomous_2023}.

\subsection{3D Rendering}

Characterizing the 3D geometry of an \textit{unknown} RSO with a single data feed is a challenging task. Until recently, there were few prospects to perform this task with high visual quality. However, recent work in the computer vision community discussed below has made this more feasible.

\textbf{Neural Radiance Fields (NeRFs)} \cite{mildenhall_nerf_2021} use a fully-connected neural network to learn a radiance field that can render synthetic views of 3D scenes. NeRFs map spatial locations and viewing angles represented as 5D coordinates $(x,y,z,\theta,\phi)$ to opacity and view-dependent color at all locations in space. The neural network estimates the opacities and colors, and rays are traced through them to compute the pixel colors of an observer's 2D viewframe of a 3D scene, an approach called volumetric rendering.

The NeRF process is differentiable end-to-end, so these models can be trained with standard gradient-based methods. NeRF is effective at generating high-quality views of 3D scenes, but has slow training and rendering times. It also only considers static scenes, which limits the possibility of movement in the scene.

\textbf{Accelerating NeRF.} Instant NeRF \cite{muller_instant_2022} accelerates NeRF training. It learns a multi-resolution set of feature vectors called neural graphics primitives that encode input scenes more efficiently. This enables Instant NeRF to produce similar results for NeRF with a much smaller neural network that trains with a far smaller computational footprint, but the rendering costs are similar to NeRF.

3D Gaussian Splatting \cite{kerbl_3d_2023} is a recent method that learns a Gaussian point-based representation of a 3D scene slightly faster than Instant NeRF but renders 1080p ($1920\times 1080$ resolution) 2D views of the scene at just 5\% of the computational cost. This method uses images of the 3D scene calibrated by Structure-from-Motion (SfM) \cite{schonberger_structure--motion_2016}. SfM computes sparse cloud of points identified in 3D based on multiple images, where 3D Gaussian random variables are initialized before being trained to represent the 3D scene.  Replacing NeRF representation with a point-based representation reduces training times since there is no huge neural network to optimize. In addition, the point-based scene representation avoids NeRF's reliance on a full volumetric representation of the scene. Hence, 3D Splatting avoids NeRF's expensive sampling of points on the paths traced by each ray and passing them through a large neural network.

The 3D Gaussian representation instead uses ``splats'' or layers of simple textures (``image primitives'') to be mixed by the Gaussian densities at different locations in space. Novel views are rendered using separate tiles of the whole screen. GPU-accelerated sorting algorithms allow extremely fast mixing for generating tiles by only considering splats that will be visible.


\textbf{Dynamic Scenes.} Dynamic NeRF (D-NeRF) \cite{pumarola_dnerf_2020} reconstructs and render novel images of objects under rigid and non-rigid motions from a single camera moving around the scene. It considers time as an additional input and splits the learning process of the fully connected network into two parts: one encodes the scene into a canonical space and one maps this canonical representation into the deformed scene at a particular time. The canonical network is trained to take in 3D coordinates and the viewing directions to yield a radiance field. Then, the deformation network outputs the displacement that takes the point to its position in the canonical space. D-NeRF can render dynamic scenes and retains high-quality image details. While D-NeRF performs better with dynamic scenes, it still faces slow training and rendering difficulties.

4D Gaussian Splatting \cite{wu_4dsplat_2023} propose a representation that uses the 3D Gaussian Splatting point-based representation and 4D neural voxels. Similar to D-NeRF, the framework is split into a 3D scene representation and a deformation field parts. However, the two pieces are time-dependent 3D Gaussian representations that can be rendered using a differential splatting process. 4D Gaussian Splatting is reported to produce high-quality dynamic scenes with the faster training and rendering than D-NeRF.

\section{Methods}

This section discusses the implementations of 3D rendering, as well as the experimental setup, datasets, and performance metrics, used in this work.

\subsection{3D Gaussian Splatting}

3D Gaussian Splatting has three parts: (1) Structure-from-Motion (SfM) learns using 2D ground truth images to calibrate the camera views of a 3D scene by learning a common point cloud representation, (2) the SfM point cloud is repurposed to initialize a 3D Gaussian point-based representation of the 3D scene, and (3) uses splatting techniques to render novel 2D views of the scene. We discuss these three steps in details below.

\textbf{Camera Calibration with SfM.} SfM reconstructs the 3D structure of the target. This involves estimating camera parameters and creating a three-dimensional point cloud for the target. We use COLMAP \cite{schonberger_structure--motion_2016} to perform the spare reconstruction of the data images. It first performs a feature detection and extraction from the presented images and then COLMAP extracts points of interest and certain features from the images that will be used in the SfM algorithm. Then, a geometric verification is performed to ensure that the images see the same scene are overlapping. Since the first matching portion does this visually, it is necessary to mathematically verify that the matched features will map to the same point in the scene. We implement COLMAP with the standard pinhole camera model.

COLMAP cycles through the data given and estimates camera parameters and poses for the images. The triangulation process can extend the set of scene points which increases coverage. A new point can be triangulated once there is another image of the scene from a different viewing angle. This can provide more correspondences between the 2D images and 3D reconstruction. It produces a point cloud as a byproduct.

\textbf{3D Gaussian Point-based Representation.} 3D Gaussisan Splatting repurposes the SfM point cloud to initialize the means of a set of 3D Gaussian random variables. Each Gaussian can be defined by its mean $p\in\mathbb{R}^3$ and covariance matrix $\Sigma\in\mathbb{R}^{3\times 3}$, with probability density
\begin{align}
    f(x)=\frac{1}{(2\pi)^{\frac{k}{2}}\sqrt{|\Sigma|}}e^{-\frac{1}{2}(x-p)^T\Sigma^{-1}(x-p)}\label{GaussianDensity}
\end{align}
The means are initialized to the positions of the SfM points and covariances initialized to isotropic covariances (diagonal matrices) based on the 3 nearest SfM-point neighbors. In addition, each Gaussian is equipped with a multiplier $\alpha\in[0,1]$ representing the opacity for capturing radiated light and spherical harmonics coefficients representing the viewing-angle-dependent color of each Gaussisan. This set of Gaussians with their means, covariances, opacities, and spherical harmonic coefficients are optimized to learn a full representation of the 3D scene using only a discrete set of points.

\textbf{3D Gaussian Splatting-based Rendering.} Splatting uses graphics primitives (e.g., textures) that extend outward from individual points to ellipsoids. The 3D Gaussian densities within these ellipsoids allow Gaussian mixing with the $\alpha$ opacities (known as $\alpha$-blending in the computer graphics community). Then, 2D views can be generated with these more complex image primitives rather than simply color and opacity in NeRF's full volumetric rendering.

The spatial extent of the splats allows point-based rendering to render equivalent or higher quality filled-in geometry without volumetric ray tracing. 3D Gaussian Splatting uses a tile-based rasterizer that pre-sorts the Gaussian splats using GPU radix search for the entire image. This allows fast tile-based $\alpha$-blending without per-pixel rendering of points. This procedure generates the required 2D views.

\textbf{Optimizing the 3D Representation}. After SfM, the entire 3D Gaussian Splatting process, both training the representation and 3D rendering, is differentiable end-to-end. This enables backpropagation and gradient-based optimization. The sparse point-based rendering faces some additional challenges in terms of how many points to use, so the method additionally has a stage where it creates or destroys Gaussians during optimization process of the Gaussian means, covariances, and associated opacities and spherical harmonic coefficients. The Gaussian covariance matrices cannot be treated as unconstrained parameters, as this would violate the definition of covariance, and hence the technique constrains the covariances to ensure the level sets of the probability densities shown in \eqref{GaussianDensity} remain ellipsoids. 

The loss function used for optimizing the scene representation and renders is as follows.

\begin{align}
(1 - \lambda) \mathcal{L}_1 + \lambda \mathcal{L}_{\text{D-SSIM}}.
\end{align}

Minimizing the first term minimizes the pixel difference between between 2D renders and ground truth images from the same viewing positions and angles. Minimizing the second term aims to increase SSIM between the images. We adopt the authors' use of $\lambda=0.2$. 

\subsection{Experiment Setup}

\begin{figure}[H]
    \centering
        \includegraphics[width=0.75\textwidth]{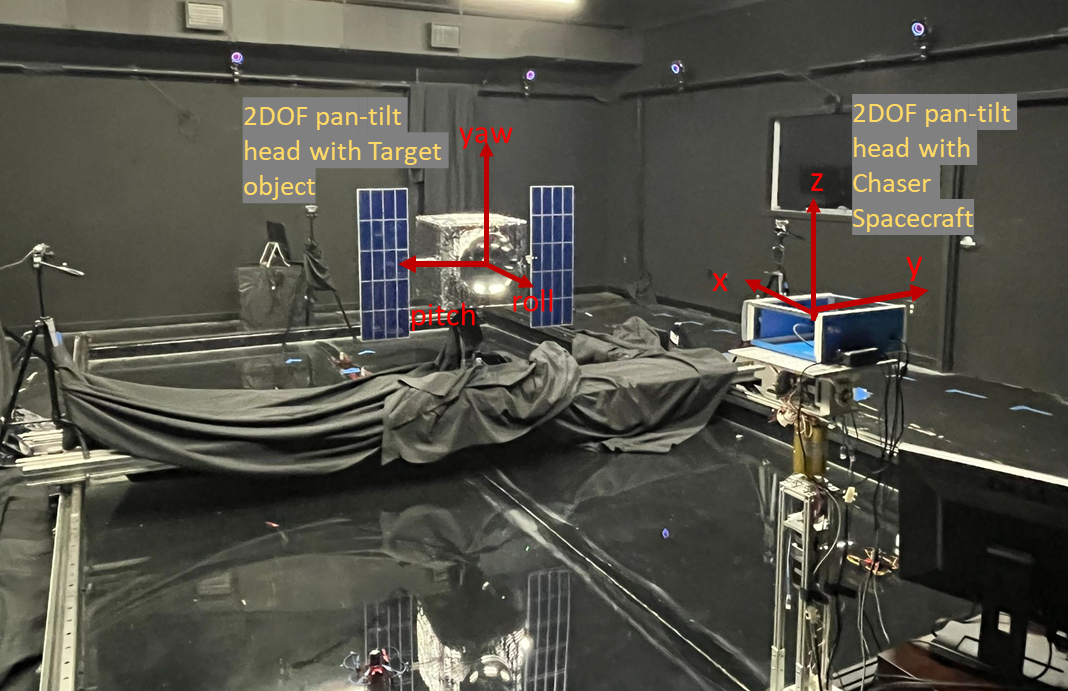}
    \caption{ORION Testbed at Florida Tech}\label{fig:ORION Testbed}
\end{figure}

HIL data was captured on the ORION testbed \cite{wilde_orion_2016} at the Autonomy Lab at Florida Tech. The test bed consists of a planar, Cartesian Maneuver kinematics simulator shown in Figure~\ref{fig:ORION Testbed} with a workspace of 5.5m x 3.5m. The simulator is equipped with a 2 DOF motion table that can translate at a maximum speed of 0.25 m/s with a maximum acceleration of 1 m/s$^{2}$. The simulator is equipped with two custom designed pan-tilt mechanisms of which both of them are capable of rotating infinitely in the azimuth direction (yawing), and rotate in elevation by $\pm 90^\circ$. In addition, one of the pan-tilt mechanisms (target object) can translate in the $x$ and $y$ axes. The target and the chaser spacecraft can be interchanged as needed. The target mock-up satellite used for this work has typical components found on a satellite such as, rectangular solar arrays, a distinct body, thruster nozzles and a parabolic antenna. 

To limit external lighting conditions, the doors, walls and floor of the testbed are painted with low reflectivity black paint and the windows are closed with black-out blind and also covered with black-out stage curtains. The lighting conditions for the experiments are simulated by a Hilio D12 LED panel with a color temperature of 5600 K (daylight balanced) rated for 350 W of power with adjustable intensity. The maximum intensity of the panel is equivalent to a 2000 W incandescent lamp. The intensity of the lamp was set to 10\% or 100\% in the scenarios described in the next section.

\subsection{Datasets}

Four HIL datasets \cite{caruso_3d_2023} were captured for training and evaluating the 3D rendering models. In all cases, images were captured using an iPhone 13 from viewing angles 360$^\circ$ around the satellite with the lamp pointing directly at the mock-up. The camera aperture was positioned at a height of 3 feet 8 inches from the ground. Each image and video is scaled to resolution 640-by-480.

The four cases differ in terms of the lighting, motion conditions, and backdrop of the images. They were captured as follows:
\begin{enumerate}
    \item[\textbf{Case 1.}] Images of the target RSO are taken at 10$^\circ$ increments around a circle of radius of 2.5 m (simulating an R-bar maneuver around a stationary satellite) with 10\% lighting intensity. Viewing angles are at a 10$^\circ$ increments rotating about the vertical axis.
    \item[\textbf{Case 2.}] Images of the target RSO are taken at 10$^\circ$ increments around a circle of radius of 2.5 m (simulating an R-bar maneuver around a stationary satellite) with 100\% lighting intensity. Viewing angles are at a 10$^\circ$ increments rotating about the vertical axis.
    \item[\textbf{Case 3.}] Videos of the RSO are captured as it yaws at 10/s with the chaser positioned 5 ft away (simulating V-bar station keeping around a spinning RSO) with 10\% lighting intensity. Viewing angles are at a 5$^\circ$ increments rotating about the vertical axis.
    \item[\textbf{Case 4.}] Videos of the RSO are captured as it yaws at 10/s with the chaser positioned 5 ft away (simulating V-bar station keeping around a spinning RSO) with 100\% lighting intensity. Viewing angles are at a 5$^\circ$ increments rotating about the vertical axis.
\end{enumerate}

\begin{figure}[H]
    \centering
    \begin{subfigure}[t]{0.49\textwidth}
        \centering
        \includegraphics[width=\textwidth]{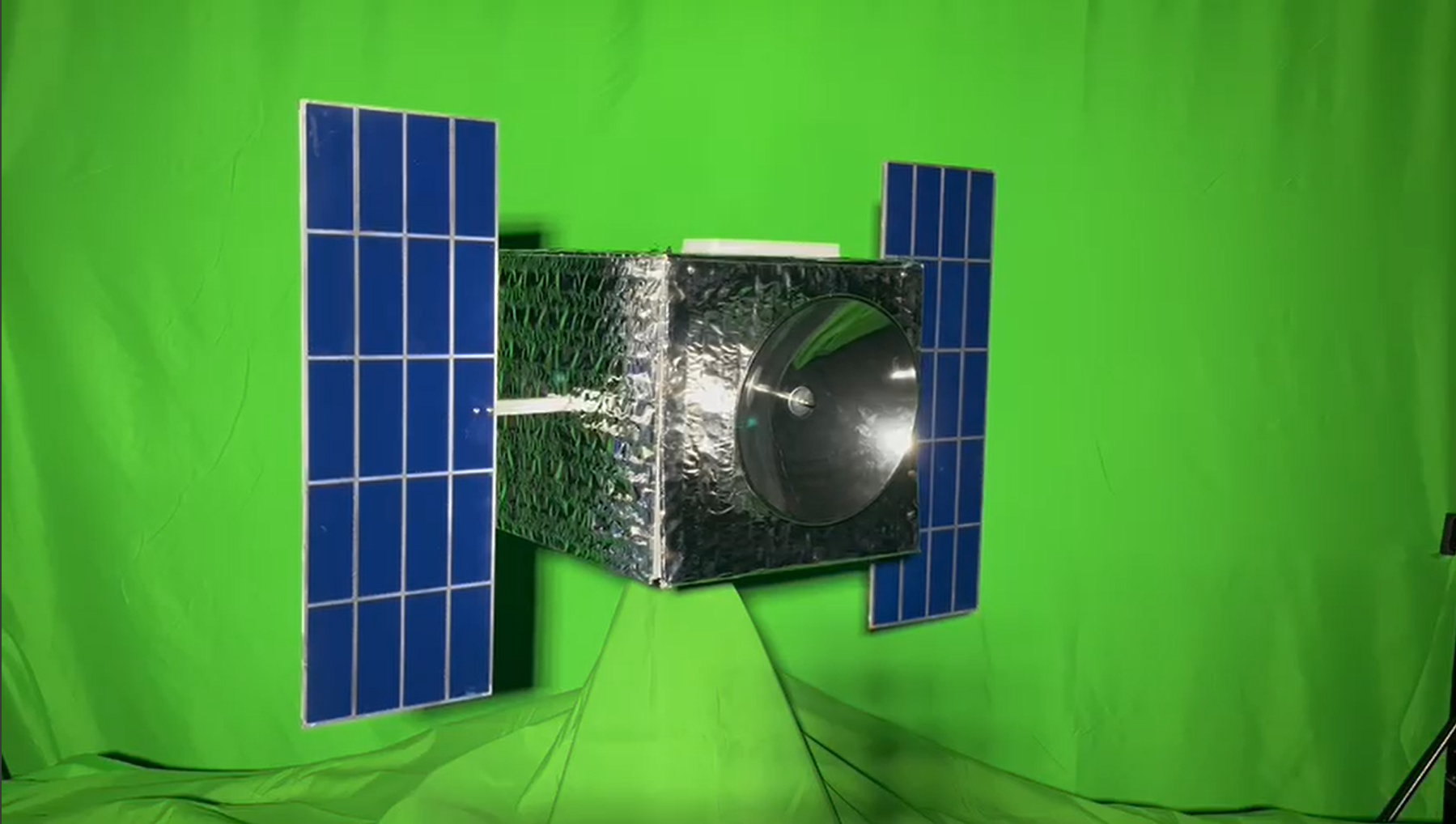}
    \end{subfigure}%
    ~
    \begin{subfigure}[t]{0.49\textwidth}
        \centering
        \includegraphics[width=\textwidth]{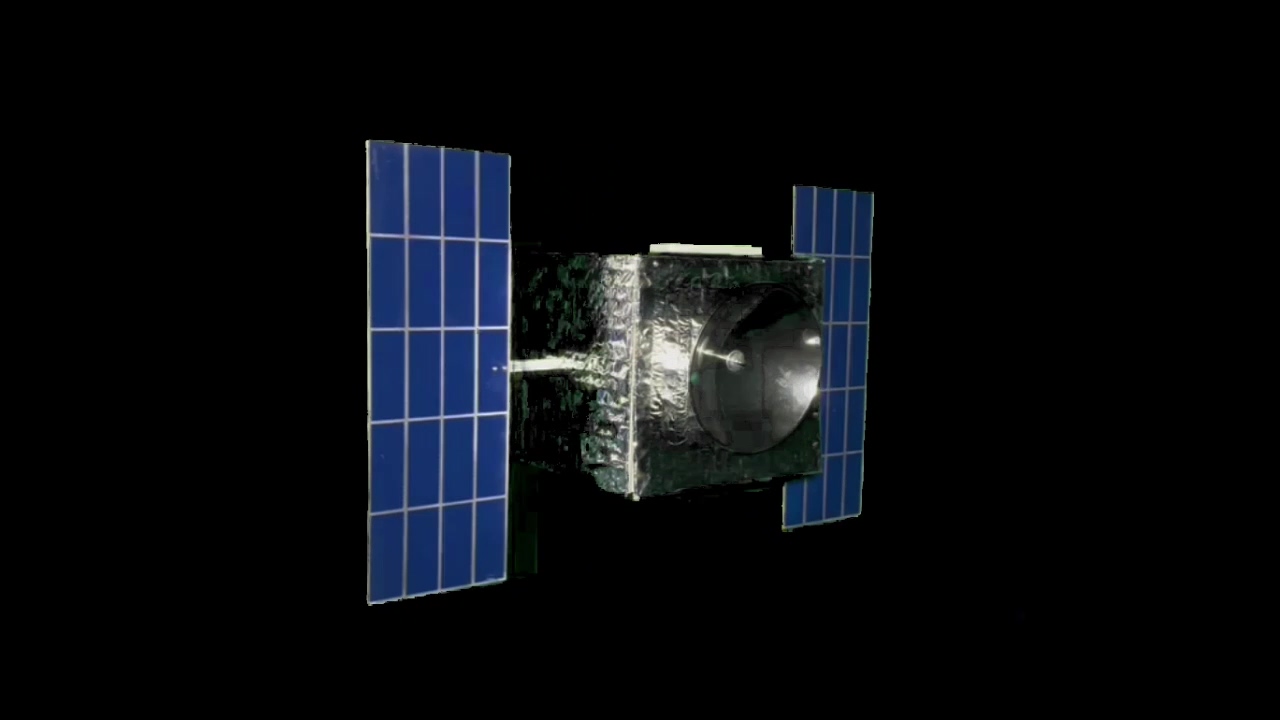}
    \end{subfigure}
    \caption{Satellite mock-up with green screen (left) and after chroma key compsiting post-processing (right).}\label{ckc}
\end{figure}

In cases 3-4, we placed a green screen behind the satellite mock-up. Using chroma key compositing, we then post-processed the images by replacing all green pixels with black. The results before and after pre-processing can be seen in Figure~\ref{ckc}.

\begin{figure}[H]
  \centering
  \begin{subfigure}{0.23\textwidth}
    \includegraphics[width=\linewidth]{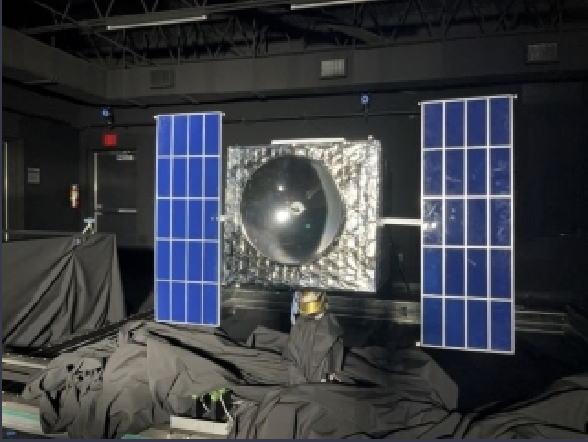}
    \subcaption{Case 1}
  \end{subfigure}
  \hfill
  \begin{subfigure}{0.23\textwidth}
    \includegraphics[width=\linewidth]{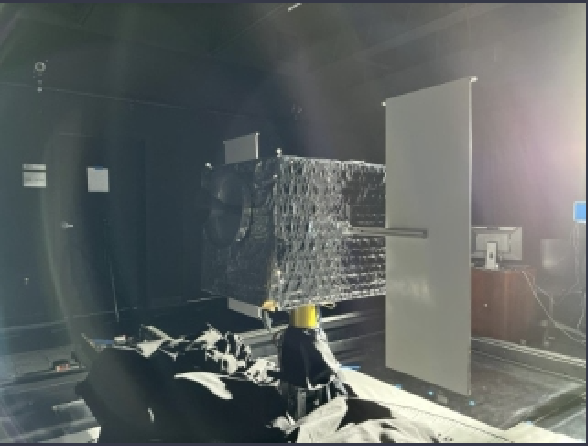}
    \subcaption{Case 2}
  \end{subfigure}
  \hfill
  \begin{subfigure}{0.23\textwidth}
    \includegraphics[width=\linewidth]{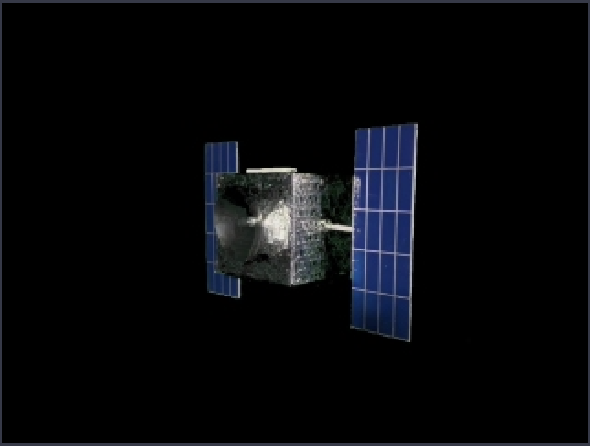}
    \subcaption{Case 3}
  \end{subfigure}
  \hfill
  \begin{subfigure}{0.23\textwidth}
    \includegraphics[width=\linewidth]{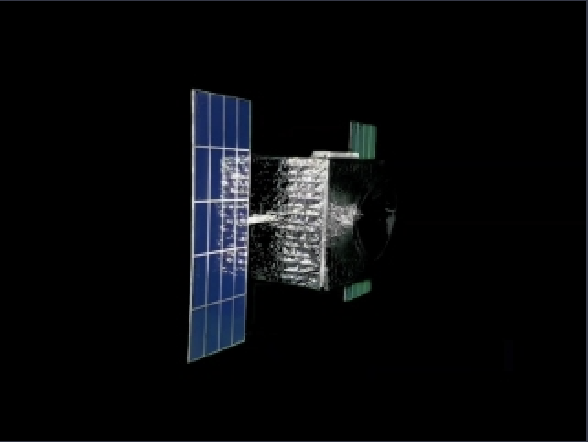}
    \subcaption{Case 4}
  \end{subfigure}
  \caption{Ground truth images of the satellite mock-up}
  \label{labDataImages}
\end{figure}

We display images from each of the four cases in Figure~\ref{labDataImages}.

\subsection{Performance Evaluation for 3D Rendering}

We measure the performance of the 3D rendering models trained herein in terms of both rending quality and computational requirements. In this subsection, we describe the metrics used.

 The quality of the renders generated by of each model trained in this work is evaluated by qualitative comparisons of individual renders as well as standard metrics for generative modeling, including Structural Similarity Index (SSIM) \cite{Wang_2004}, Peak Signal to Noise Ratio (PSNR), and Learned Perceptual Image Patch Similarity (LPIPS) \cite{zhang_unreasonable_2018} evaluated on images of the satellite mock-up not used during training.

SSIM measures the difference between the original picture and the reconstructed image that accounts for changes such as luminance or contrast. High SSIM values indicate high performance. PSNR measures the quality of rendered images at the pixel level. High PSNR indicates good performance. LPIPS is a more advanced metric that aims to capture human-perceived similarity of images. It feeds both a real and a synthetic image patch to a VGG network \cite{simonyan_very_2015} trained to classify the ImageNet dataset \cite{ILSVRC15}, extracts the hidden representations of each patch within different layers of VGG, and computes a distance between them. This distance is fed to a model to mimic human preferences, yielding the LPIPS score. Lower LPIPS scores indicate the synthetic images are more similar to real images.

In addition, methods are evaluated in terms of computational requirements during both training and inference (i.e., 3D rendering). For training, we compare VRAM usage and training time. For inference (i.e. rendering), we compare VRAM usage and framerate (frames per second - FPS). VRAM usage is measured with \texttt{nvidia-smi pmon} monitoring utility every 1 second, using the highest number logged, when all data and the current model are fully loaded in GPU. For FPS measurement. Due to extra time taken for data and model copy from CPU to GPU, first couple of rendering runs will be slower until data are fully cached. Hence, we measured FPS by looping render task on the test set for 10-20 iterations, then take the average of the median 5 runs.

\section{Results}

In this section, we present 3D rendering results and performance by 3D Gaussian Splatting (3DGS) on all test datasets and compared with several competing methods, including NeRF \cite{mildenhall_nerf_2021}, D-NeRF \cite{pumarola_dnerf_2020}, Instant NeRF \cite{muller_instant_2022}, and 4D Gaussian Splatting (4DGS) \cite{wu_4dsplat_2023}.

The new results differ from our previous work \cite{caruso_3d_2023} because that paper limited its quantitative analysis to images in the training set while novel renders were analyzed qualitatively.

\begin{table}[H]
\centering
\caption{Quality of 3D Renders by Models and different Test Cases}
\label{tab:cases-methods}
\resizebox{1.0\textwidth}{!}{
\begin{tabular}{lcccccccccccc}
\toprule
        & \multicolumn{3}{c}{\textbf{Case 1}} 
        & \multicolumn{3}{c}{\textbf{Case 2}} 
        & \multicolumn{3}{c}{\textbf{Case 3}} 
        & \multicolumn{3}{c}{\textbf{Case 4}} \\
\cmidrule(lr){2-4} \cmidrule(lr){5-7} \cmidrule(lr){8-10} \cmidrule(lr){11-13}
\textbf{Method}  
    & \textbf{SSIM$\uparrow$} & \textbf{PSNR$\uparrow$} & \textbf{LPIPS$\downarrow$} 
    & \textbf{SSIM$\uparrow$} & \textbf{PSNR$\uparrow$} & \textbf{LPIPS$\downarrow$} 
    & \textbf{SSIM$\uparrow$} & \textbf{PSNR$\uparrow$} & \textbf{LPIPS$\downarrow$} 
    & \textbf{SSIM$\uparrow$} & \textbf{PSNR$\uparrow$} & \textbf{LPIPS$\downarrow$} \\
\midrule
NeRF 
    & 0.3891  & 16.81  & 0.5503   
    & 0.4520  & 16.52  & 0.5008   
    & \textbf{0.9332}  & \textbf{28.24}  & \textbf{0.0763}   
    & 0.6438  & 19.54  & 0.3207  \\
D-NeRF     
    & 0.3783  & 16.76  & 0.5418   
    & 0.0337  & 9.22   & 0.5877   
    & 0.8156  & 16.66  & 0.1853   
    & 0.6184  & 19.61  & 0.3282  \\
Instant NeRF     
    & 0.5149  & 16.47  & 0.4374   
    & 0.4729  & 14.71  & 0.4440   
    & 0.8571  & 17.55  & 0.1277   
    & 0.8569  & 19.99  & 0.1056  \\
3DGS
    & \textbf{0.9223}    & \textbf{25.70}    & \textbf{0.0814}     
    & \textbf{0.6803}    & \textbf{16.78}    & \textbf{0.2949}     
    & 0.8756    & 26.53  & 0.1040
    & \textbf{0.9213}    & \textbf{25.52}    & \textbf{0.0796}    \\
4DGS     
    & 0.5192  & 16.78  & 0.4310   
    & 0.4877  & 13.17  & 0.5193  
    & 0.4358  & 15.07  & 0.1454   
    & 0.7619  & 17.16  & 0.1890  \\

\bottomrule
\end{tabular}}
\end{table}

At a first glance, a previously discussed trend (in \cite{caruso_3d_2023} reappeared in our new result regardless of the slight differences in methodology: Dynamic Scene (D-NeRF, 4DGS) reconstruction methods took longer to train (Table \ref{tab:cases-methods-memory}) and consistently perform worse (Table \ref{tab:cases-methods}) than their static counterparts (NeRF, Instant Nerf, 3DGS). In addition, training VRAM usage metrics shows these dynamic scene models also consume more resources. As such, our previous conclusion still stands: static scene reconstruction methods are better for 3D reconstruction of satellites.

The extreme lighting case with background noise (Case 2) still poses some difficulties for all models \cite{caruso_3d_2023}, with all metrics being consistently worse than the same case with lower lighting (Case 1). Interestingly, when removing the background in post-processing (Case 3 and Case 4), increased light exposure does not negatively impact the performance of Instant NeRF or 3DGS. Even with 4DGS, the performance degradation is not as severe as typical NeRF-based models. This shows that Gaussian Splatting-based models (and Instant NeRF) are the best models to reconstruct models under extreme lighting conditions in space.

Next, we focus on computing and memory requirements of the five algorithms trained on the four test cases. All are compared using a single NVIDIA RTX 3080 Ti. Training times and rendering times will be higher on spaceflight hardware, but it is feasible for some of the methods using, e.g., an NVIDIA Jetson.

\begin{table}[H]
\centering
\caption{Model Training Computational Requirements}
\label{tab:cases-methods-memory}
\resizebox{1.0\textwidth}{!}{
\begin{tabular}{lcccccccccccc}
\toprule
        & \multicolumn{2}{c}{\textbf{Case 1}} 
        & \multicolumn{2}{c}{\textbf{Case 2}} 
        & \multicolumn{2}{c}{\textbf{Case 3}} 
        & \multicolumn{2}{c}{\textbf{Case 4}} \\
\cmidrule(lr){2-3} \cmidrule(lr){4-5} \cmidrule(lr){6-7} \cmidrule(lr){8-9}
\textbf{Method}  
    & \textbf{VRAM (MB)$\downarrow$} & \textbf{Training time$\downarrow$} 
    & \textbf{VRAM (MB)$\downarrow$} & \textbf{Training time$\downarrow$} 
    & \textbf{VRAM (MB)$\downarrow$} & \textbf{Training time$\downarrow$}
    & \textbf{VRAM (MB)$\downarrow$} & \textbf{Training time$\downarrow$} \\
\midrule
NeRF 
    & 3833  & 54m24s
    & 3845  & 55m47s  
    & 4195  & 1h18m21s   
    & 3952  & 1h15m24s \\
D-NeRF     
    & 4586  & 1h37m17s
    & 4598  & 1h39m32s  
    & 4948  & 2h14m56s   
    & 4634  & 2h05m24s \\
Instant NeRF
    & 1664  & \textbf{5m18s}
    & 1684  & \textbf{5m7s}  
    & 1934  & \textbf{6m11s}   
    & \textbf{1702}  & \textbf{6m15s} \\
3DGS
    & \textbf{1485}  & 5m42s
    & \textbf{1668}  & 7m4s  
    & \textbf{1875}  & 6m59s   
    & 1792  & 6m32s \\
4DGS     
    & 2224  & 57m32s
    & 4089  & 1h5m51s  
    & 5385  & 44m34s   
    & 5022  & 1h56m00s \\
\bottomrule
\end{tabular}}
\end{table}

Training time and resource consumption is shown in Table~\ref{tab:cases-methods-memory}. All NeRF-based models stay consistent across all lighting cases, which is due to the neural network used for the models are the same for each method, and most of the differences are due to different amount of training/testing data for each cases being cached on GPU. On the other hand, Gaussian Splatting-based methods have different resource usage pattern across all cases. Since during training, 3D/4D Gaussians are being continuously added and removed from random initialization \cite{kerbl_3d_2023,wu_4dsplat_2023}, the model size will be vastly different depends on their local optimized states and the amount of Gaussians/complexity of the scenes. 

Instant NeRF and 3DGS training resource consumption and training time are the best across all compared methods. However, regarding quality of 3D model reconstruction (see Table \ref{tab:cases-methods}) in cases with background noise (Case 1 and 2), 3DGS performs exceptionally better, while in the background removed cases (Case 3 and 4), both methods perform similarly. Original implementation of NeRF on average takes 10 times longer to train, while only performs marginally better than the previously mentioned methods in Case 3.

\begin{table}[H]
\centering
\caption{Model Inference: Computational Requirements and Rendering Framerates}
\label{tab:inference-memory}
\resizebox{1.0\textwidth}{!}{
\begin{tabular}{lcccccccccccc}
\toprule
        & \multicolumn{2}{c}{\textbf{Case 1}} 
        & \multicolumn{2}{c}{\textbf{Case 2}} 
        & \multicolumn{2}{c}{\textbf{Case 3}} 
        & \multicolumn{2}{c}{\textbf{Case 4}} \\
\cmidrule(lr){2-3} \cmidrule(lr){4-5} \cmidrule(lr){6-7} \cmidrule(lr){8-9}
\textbf{Method}  
    & \textbf{VRAM (MB)$\downarrow$} & \textbf{FPS$\uparrow$} 
    & \textbf{VRAM (MB)$\downarrow$} & \textbf{FPS$\uparrow$} 
    & \textbf{VRAM (MB)$\downarrow$} & \textbf{FPS$\uparrow$}
    & \textbf{VRAM (MB)$\downarrow$} & \textbf{FPS$\uparrow$} \\
\midrule
NeRF 
    & 9143   & 0.27
    & 11307  & 0.31  
    & 11323  & 0.14   
    & 11323  & 0.07 \\
D-NeRF     
    & 9171   & 0.19
    & 11331  & 0.20  
    & 11347  & 0.08   
    & 11349  & 0.05 \\
Instant NeRF
    & 1375  & 0.67
    & 1407  & 0.85  
    & 1829  & 0.59   
    & \textbf{1827}  & 0.23 \\
3DGS 
    & \underline{1615}  & \textbf{45.84}
    & \textbf{1129}     & \textbf{215.12}  
    & \underline{2807}  & \underline{108.78}   
    & \underline{1955}     & \textbf{42.04} \\
4DGS     
    & \textbf{1351}  & 23.23
    & 1389           & 28.93  
    & \textbf{1681}  & \textbf{121.85}   
    & 2687           & 8.23 \\
    
\bottomrule
\end{tabular}}
\end{table}

While Instant NeRF trains with impressively low resource consumption and renders 3-5 times faster than NeRF and D-NeRF, it is still vastly slower than Gaussian Splatting-based rendering. This shows benefits of the Gaussian Splatting rasterization method, with 3DGS reaching 45.84 FPS, nearly two orders of magnitude faster than even Instant NeRF.

Combining performance metrics with training time and memory usage, fast rasterization and rendering speed, we conclude that 3DGS is the best model to characterize the geometry of the RSO, especially when considering the computing limitations of spacecraft hardware.

\begin{figure}[htbp]
\centering
\begin{minipage}{0.05\textwidth}
\centering
\phantom{Cases}
\end{minipage}
\begin{minipage}{0.15\textwidth}
\centering
Ground Truth
\end{minipage}
\begin{minipage}{0.15\textwidth}
\centering
NeRF
\end{minipage}
\begin{minipage}{0.15\textwidth}
\centering
D-NeRF
\end{minipage}
\begin{minipage}{0.15\textwidth}
\centering
Instant NeRF
\end{minipage}
\begin{minipage}{0.15\textwidth}
\centering
3DGS
\end{minipage}
\begin{minipage}{0.15\textwidth}
\centering
4DGS
\end{minipage}

\begin{minipage}[t]{0.05\textwidth}
\centering
\rotatebox{90}{\phantom{xy} Case 1}
\end{minipage}
\begin{subfigure}{0.15\textwidth}
\includegraphics[width=\linewidth, angle=180, origin=c]{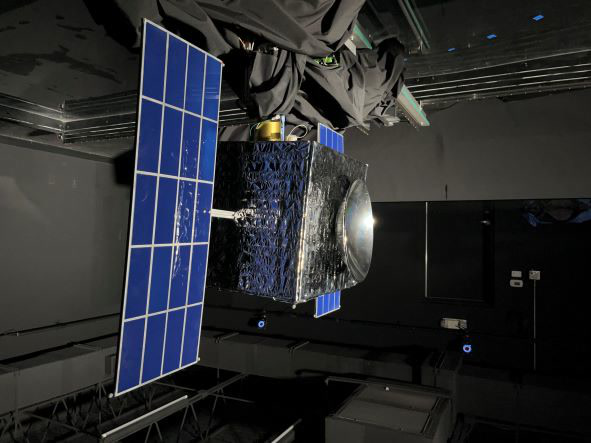}
\end{subfigure}
\begin{subfigure}{0.15\textwidth}
\includegraphics[width=\linewidth, angle=180, origin=c]{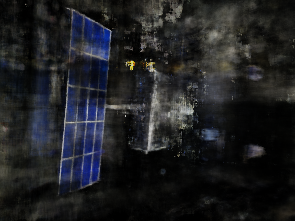}
\end{subfigure}
\begin{subfigure}{0.15\textwidth}
\includegraphics[width=\linewidth, angle=180, origin=c]{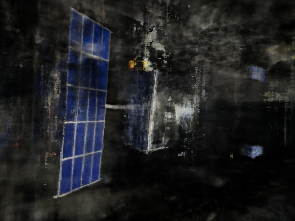}
\end{subfigure}
\begin{subfigure}{0.15\textwidth}
\includegraphics[width=\linewidth, angle=180, origin=c]{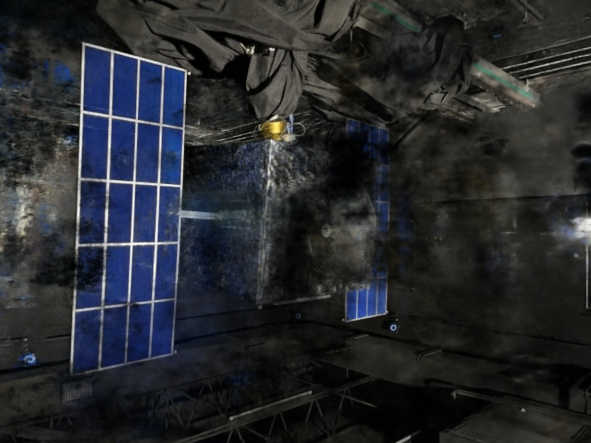}
\end{subfigure}
\begin{subfigure}{0.15\textwidth}
\includegraphics[width=\linewidth, angle=180, origin=c]{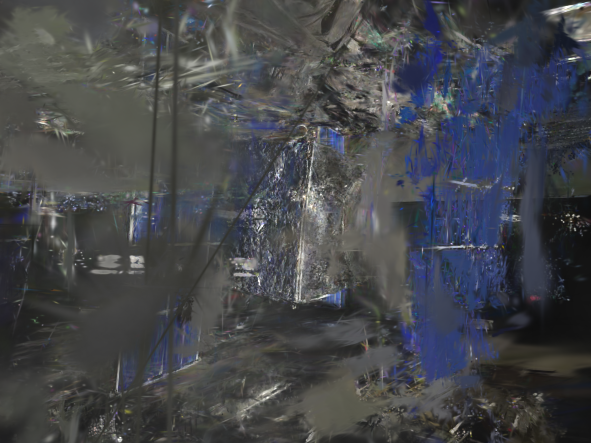}
\end{subfigure}
\begin{subfigure}{0.15\textwidth}
\includegraphics[width=\linewidth, height=0.75\linewidth, angle=180, origin=c]{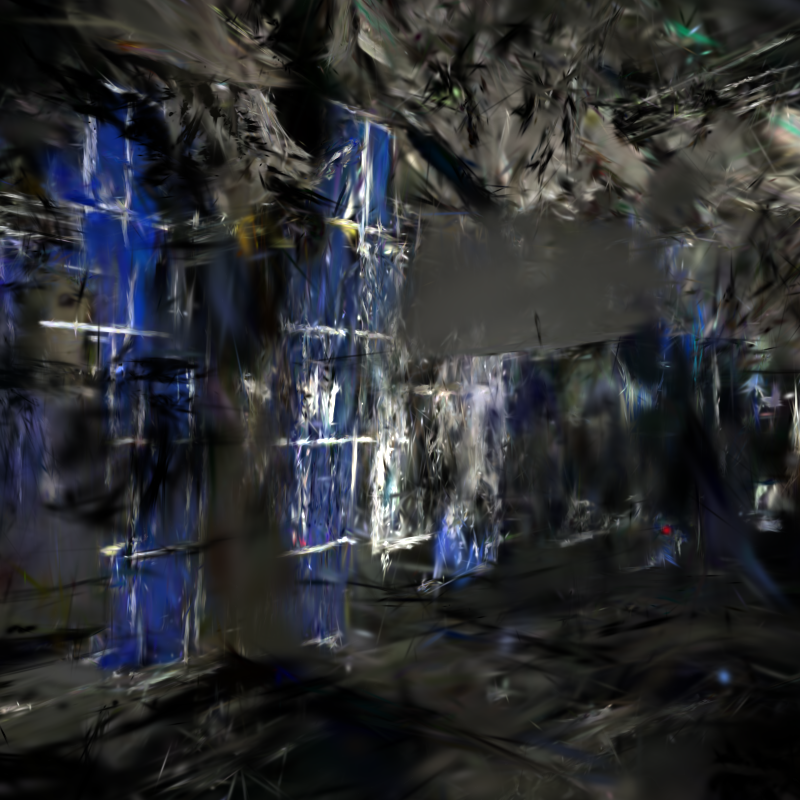}
\end{subfigure}

\begin{minipage}[t]{0.05\textwidth}
\centering 
\rotatebox{90}{\phantom{xy} Case 2}
\end{minipage}
\begin{subfigure}{0.15\textwidth}
\includegraphics[width=\linewidth]{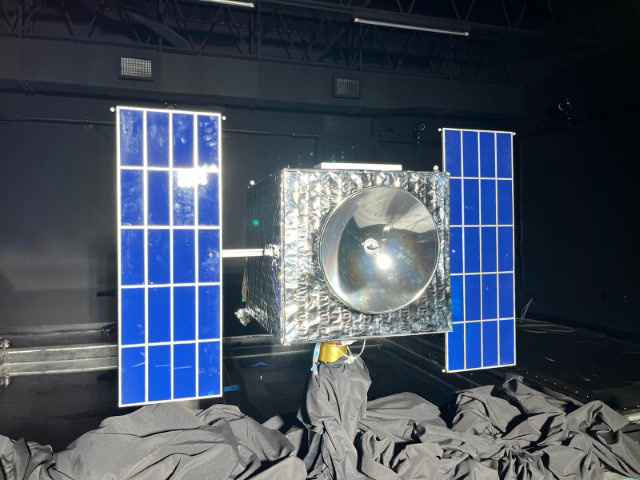}
\end{subfigure}
\begin{subfigure}{0.15\textwidth}
\includegraphics[width=\linewidth]{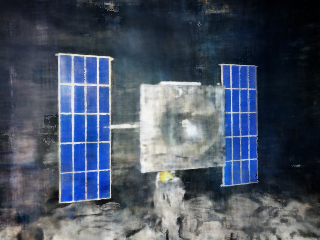}
\end{subfigure}
\begin{subfigure}{0.15\textwidth}
\includegraphics[width=\linewidth]{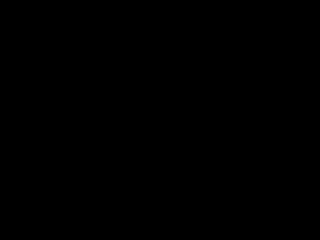}
\end{subfigure}
\begin{subfigure}{0.15\textwidth}
\scalebox{-1}[1]{\includegraphics[width=\linewidth]{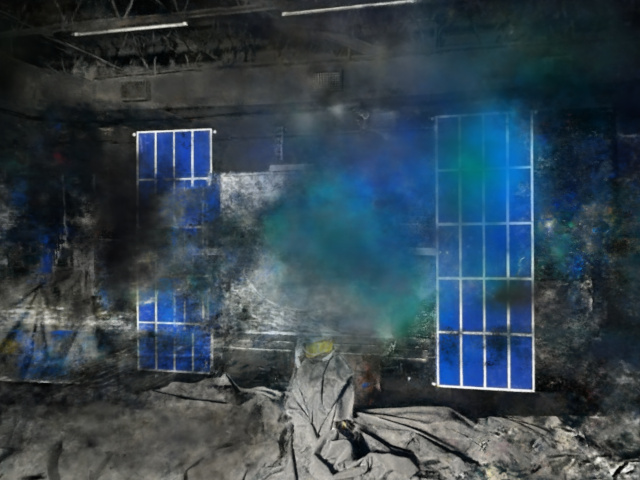}}
\end{subfigure}
\begin{subfigure}{0.15\textwidth}
\includegraphics[width=\linewidth]{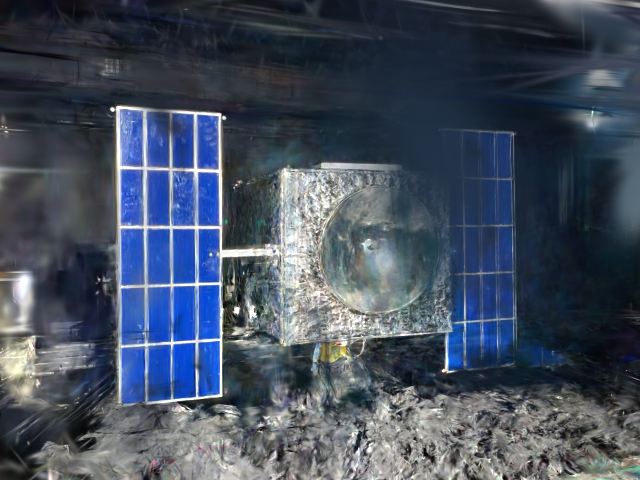}
\end{subfigure}
\begin{subfigure}{0.15\textwidth}
\includegraphics[width=\linewidth, height=0.75\linewidth]{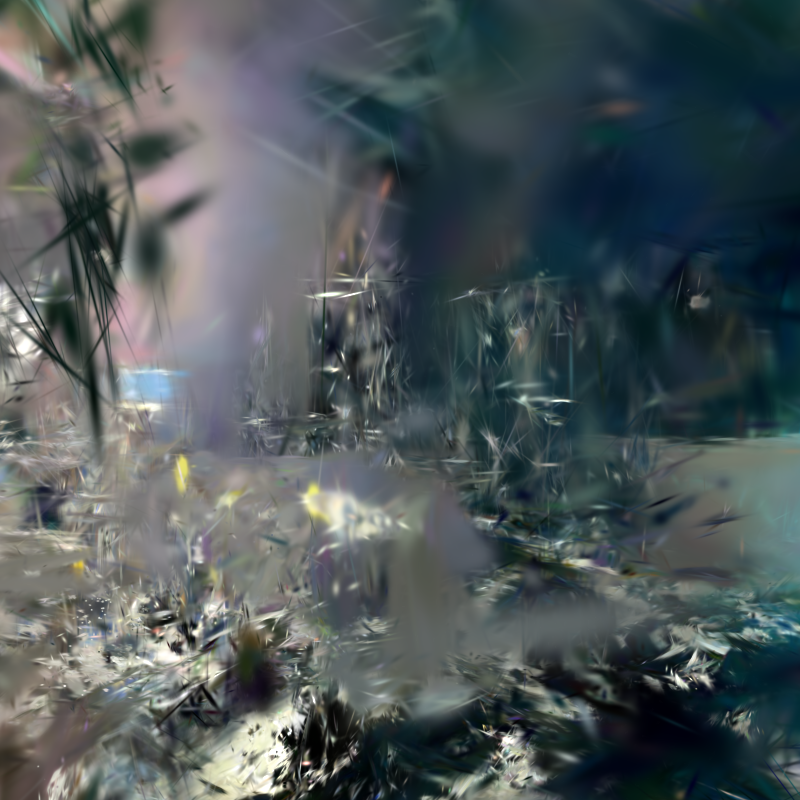}
\end{subfigure}

\begin{minipage}[t]{0.05\textwidth}
\centering 
\rotatebox{90}{\phantom{x} Case 3}
\end{minipage}
\begin{subfigure}{0.15\textwidth}
\includegraphics[width=\linewidth]{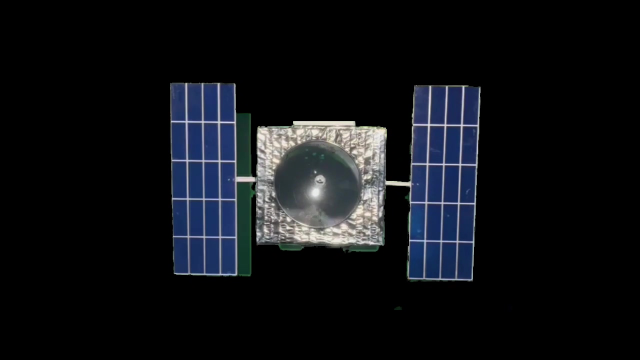}
\end{subfigure}
\begin{subfigure}{0.15\textwidth}
\includegraphics[width=\linewidth]{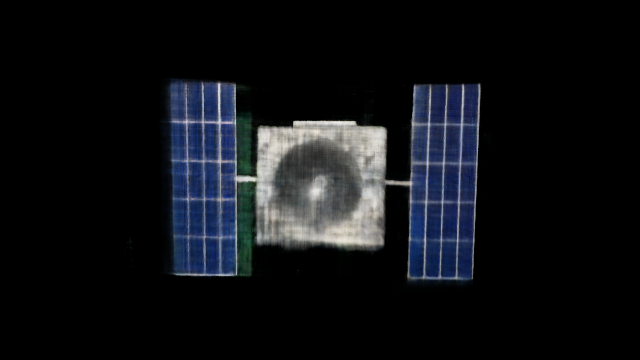}
\end{subfigure}
\begin{subfigure}{0.15\textwidth}
\includegraphics[width=\linewidth]{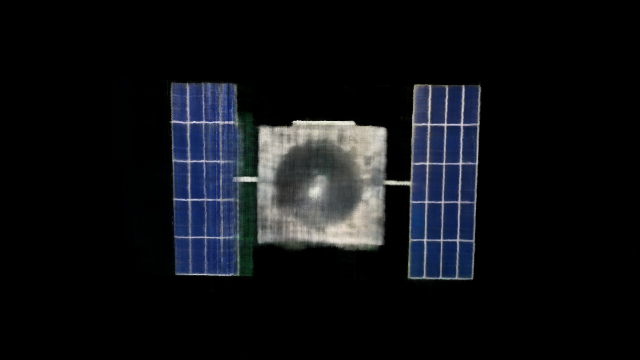}
\end{subfigure}
\begin{subfigure}{0.15\textwidth}
\includegraphics[width=\linewidth]{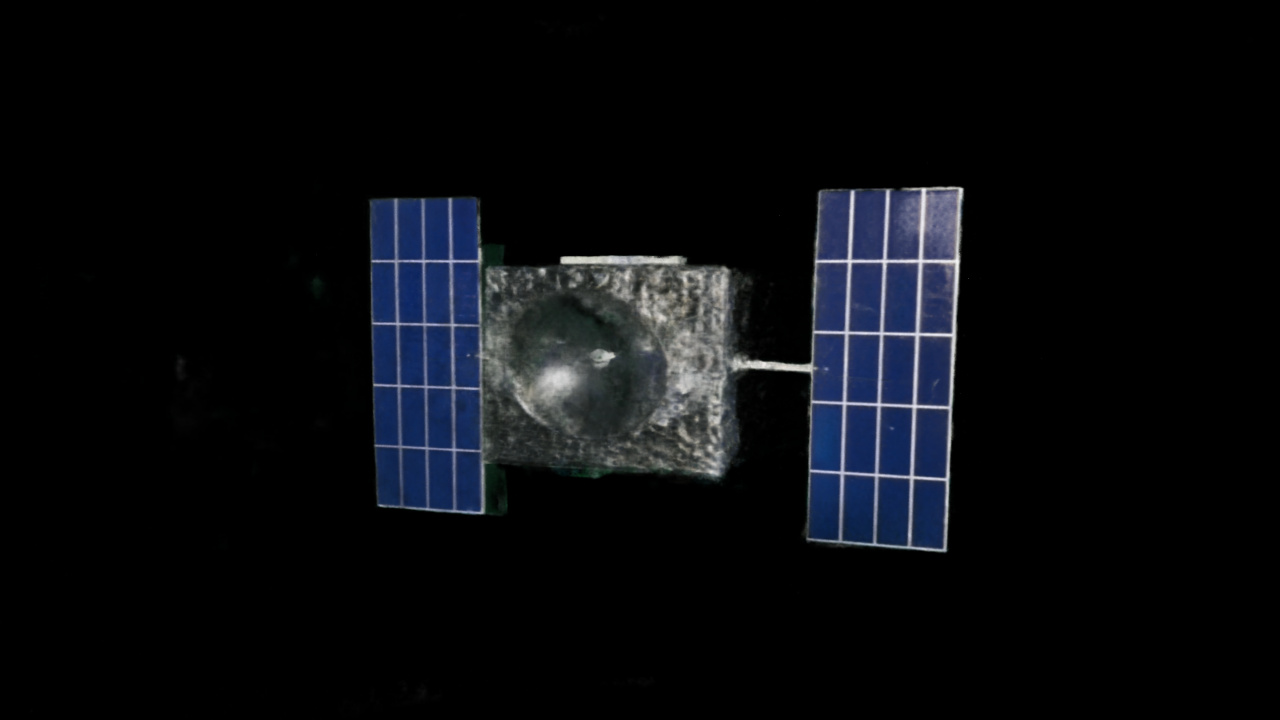}
\end{subfigure}
\begin{subfigure}{0.15\textwidth}
\includegraphics[width=\linewidth]{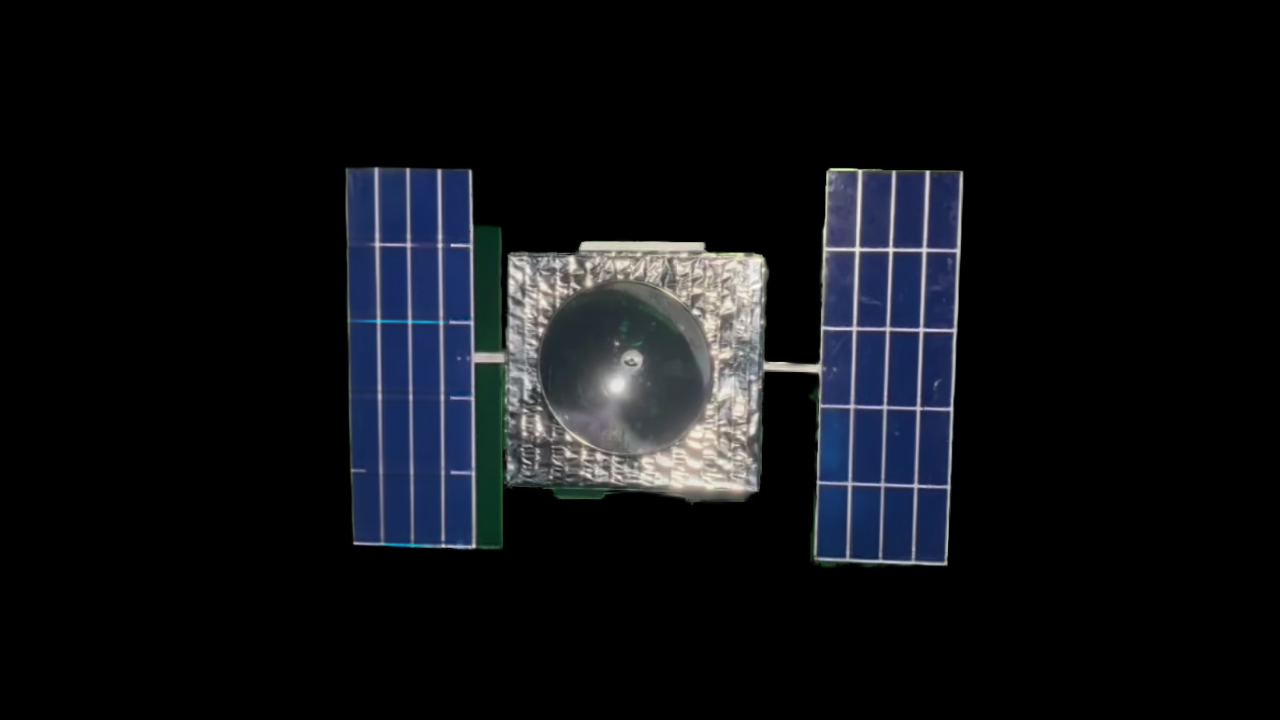}
\end{subfigure}
\begin{subfigure}{0.15\textwidth}
\includegraphics[width=\linewidth, height=0.564\linewidth]{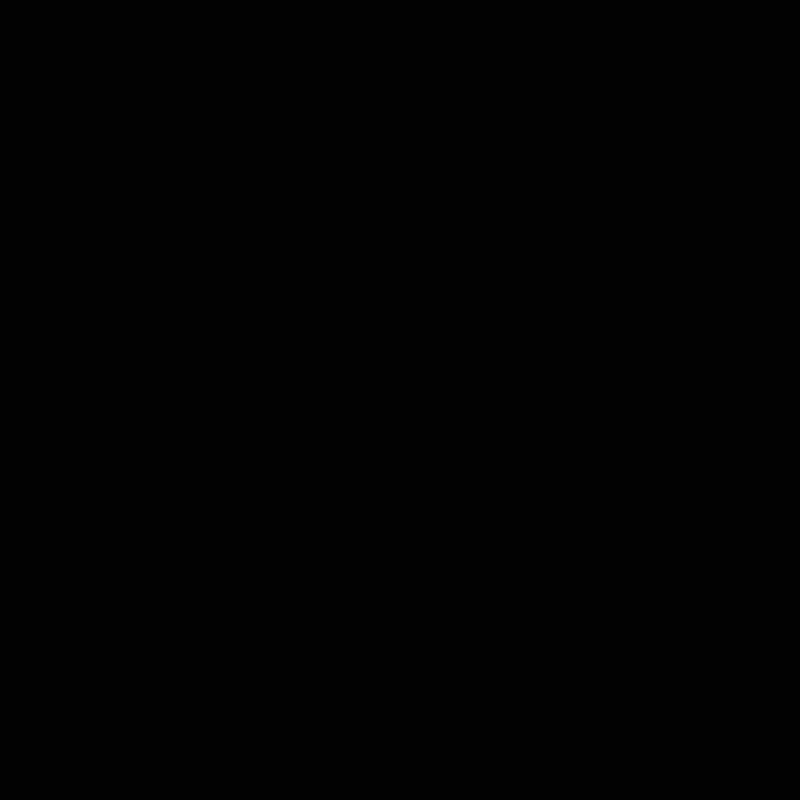}
\end{subfigure}

\begin{minipage}[t]{0.05\textwidth}
\centering 
\rotatebox{90}{\phantom{x} Case 4}
\end{minipage}
\begin{subfigure}{0.15\textwidth}
\includegraphics[width=\linewidth]{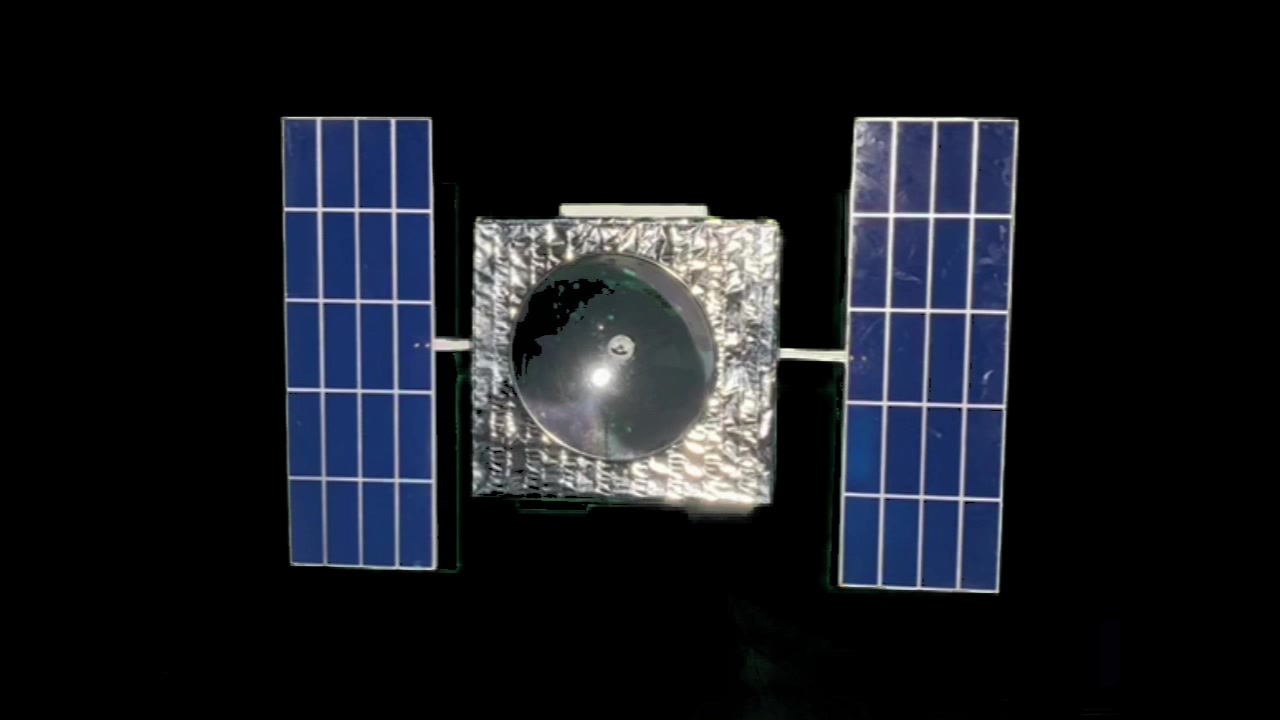}
\end{subfigure}
\begin{subfigure}{0.15\textwidth}
\includegraphics[width=\linewidth]{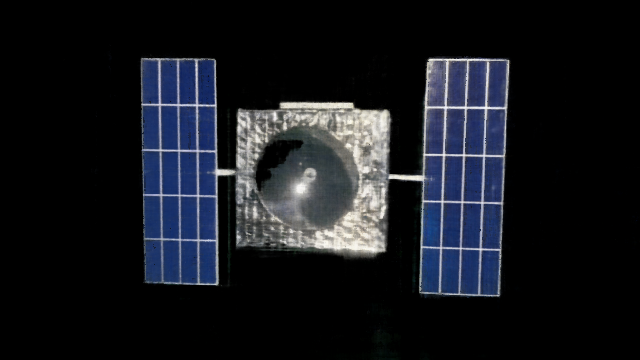}
\end{subfigure}
\begin{subfigure}{0.15\textwidth}
\includegraphics[width=\linewidth]{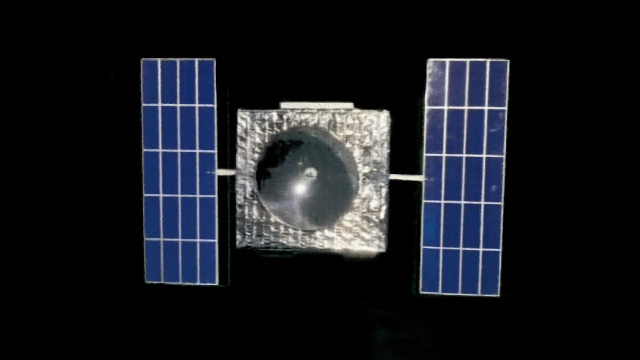}
\end{subfigure}
\begin{subfigure}{0.15\textwidth}
\includegraphics[width=\linewidth]{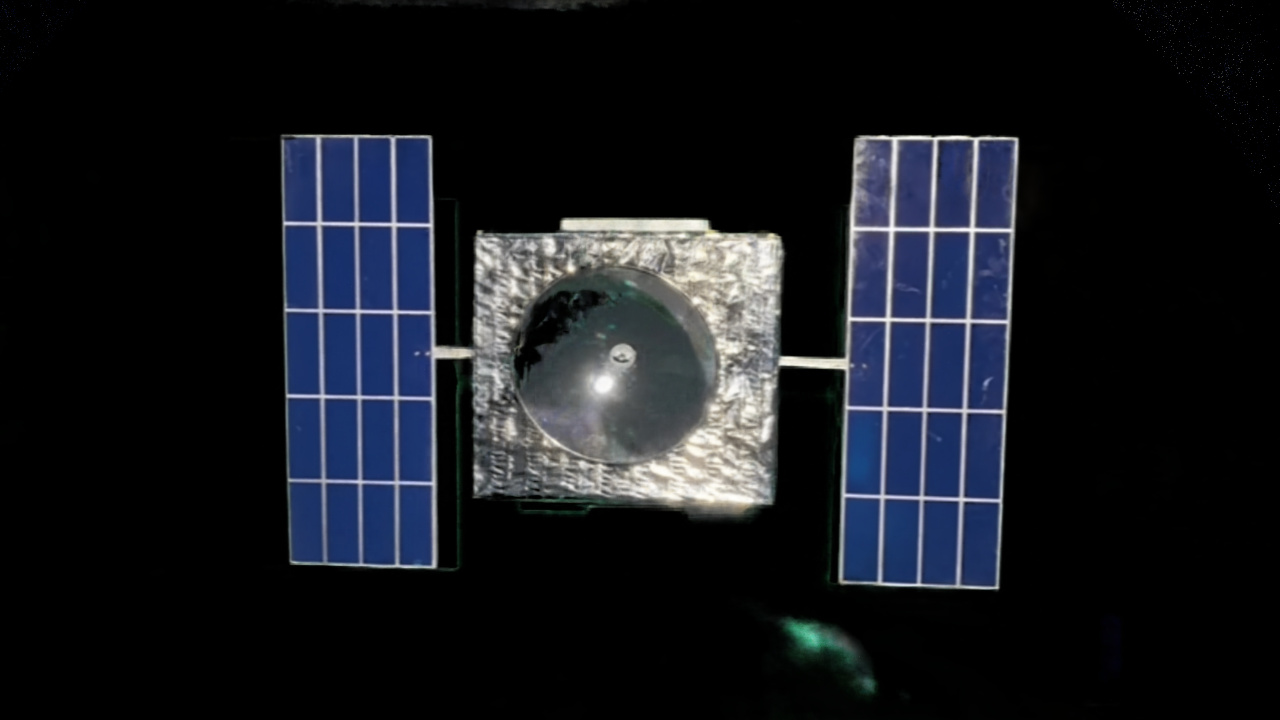}
\end{subfigure}
\begin{subfigure}{0.15\textwidth}
\includegraphics[width=\linewidth]{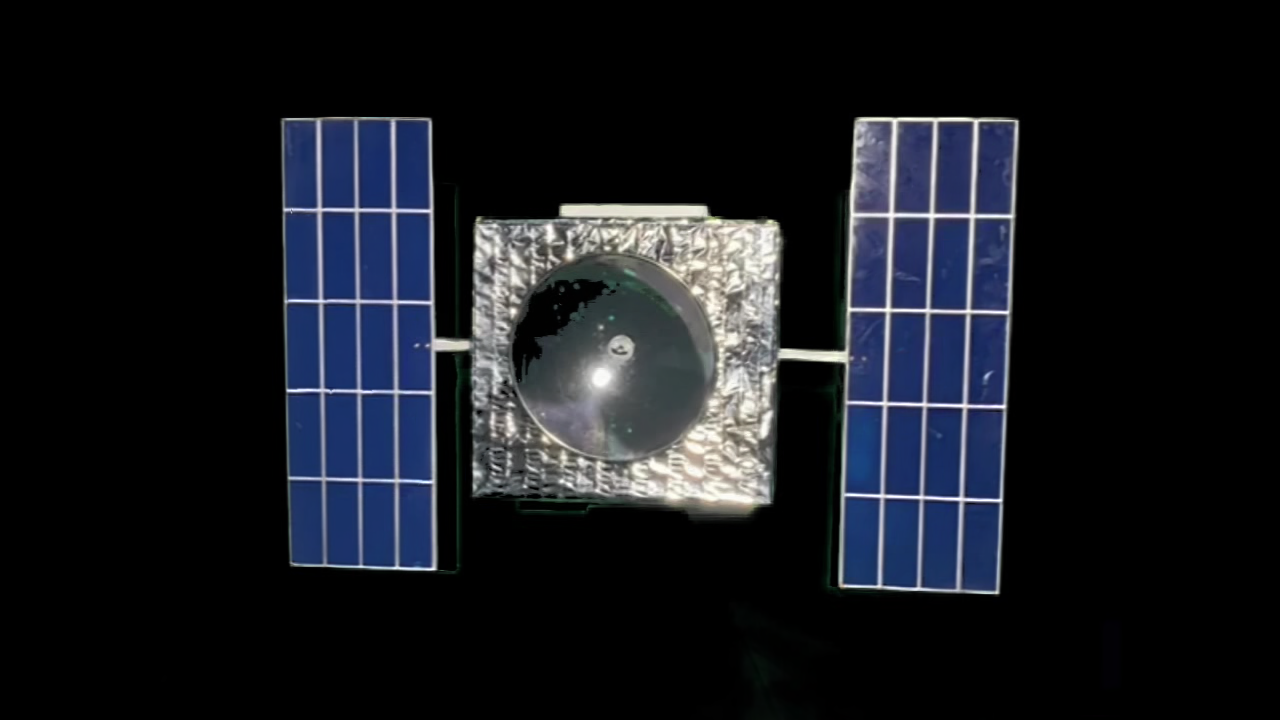}
\end{subfigure}
\begin{subfigure}{0.15\textwidth}
\includegraphics[width=\linewidth, height=0.564\linewidth]{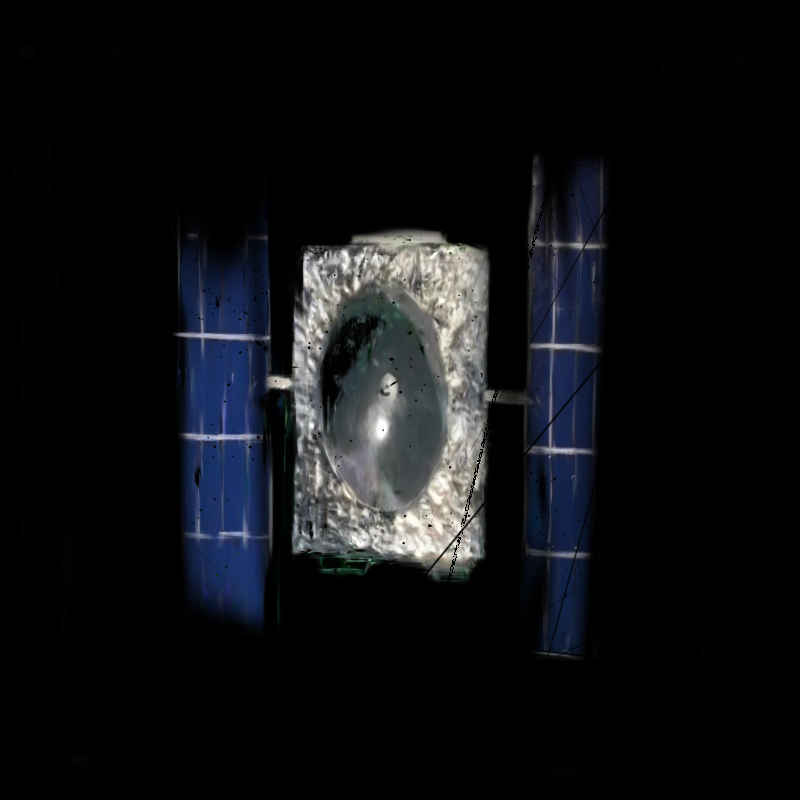}
\end{subfigure}

\caption{Comparison of different test cases and SfM methods.}
\label{fig:cases-methods}
\end{figure}

Qualitative comparison of novel view synthesis between models is shown in Figure \ref{fig:cases-methods}. Both NeRF and 3DGS can render accurate views of the satellite. In the samples presented, NeRF-rendered images have lots of blur, artifacts, and boundaries of objects that ``melt into'' on another. Although 4DS failed in the render shown for Case 1, Gaussian Splatting-based renders have less blurring and artifacts, produce sharper reflections, and preserve finer details on satellite surfaces. 

Similar to the quantitative results above, dynamic scene reconstruction methods perform poorly under extreme lighting: in particular, D-NeRF and 4DGS in Case 2. Renders that are fully black (Case 2 for D-NeRF and Case 3 for 4DGS) are not missing or failed renders. They are real renders, but they exhibit extreme occlusion artifacts that cover the entire view supplied to the model. Note the darkened extemities of the solar pannels in the Case 4 render by 4DGS for a less extreme example. Additionally, 4DGS fails to reconstruct a reliable representation of the satellite model in all cases.

\section{Conclusions}


This work demonstrates 3D Gaussian Splatting models can be trained to learn expressive 3D scene representations of unknown, non-cooperative satellites on orbit and generate high-quality 2D renders from novel views. Unlike prior work, it can be done on the limited compute capacity of current spaceflight hardware.

The demonstrates effective capabilities for on-board characterization of RSO geometries. This can serve as an important enabler wider on-board characterization of spacecraft on orbit, including component recognition and inspection, pose estimation, kinematics characterization, and other inspection tasks. These enable downstream autonomous GNC and RPO to support OOS and ADR missions.

\section*{Acknowledgments}

R. T. White thanks the NVIDIA Applied Research Accelerator Program for hardware support. 

\bibliographystyle{ieeetr}
\bibliography{splat}

\end{document}